\documentclass[12pt]{article}
\usepackage[margin=1in]{geometry}
\usepackage{amsmath,amssymb}
\usepackage{hyperref}
\usepackage{natbib}
\usepackage{setspace}
\usepackage{titlesec}
\title{\textbf{The Specification Trap:\\Why Static Value Alignment Alone Is Insufficient for Robust Alignment}}
\author{Austin Spizzirri\\Belmont University\\\texttt{austin.spizzirri@gmail.com}}
\date{April 2026}

\begin{document}
\maketitle

\begin{abstract}
Static content-based AI value alignment is insufficient for robust alignment under capability scaling, distributional shift, and increasing autonomy. This holds for any approach that treats alignment as optimizing toward a fixed formal value-object, whether reward function, utility function, constitutional principles, or learned preference representation.

Three philosophical results create compounding difficulties: Hume's is-ought gap (behavioral data underdetermines normative content), Berlin's value pluralism (human values resist consistent formalization), and the extended frame problem (any value encoding will misfit future contexts that advanced AI creates). RLHF, Constitutional AI, inverse reinforcement learning, and cooperative assistance games each instantiate this \textit{specification trap}, and their failure modes reflect structural vulnerabilities, not merely engineering limitations that better data or algorithms will resolve. The workarounds for individual components are mutually undermining: each relies on a condition that another component denies.

The trap activates at closure: the moment a specification ceases to update from the process it governs. This is the paper's central contribution. Recent empirical work confirms that the resulting gap is live rather than speculative: frontier models deployed with agentic capabilities pursue misaligned objectives at rates far exceeding supervised-evaluation baselines \citep{lynch2025}. Drawing on compatibilist philosophy, the paper argues that behavioral compliance under training conditions does not guarantee robust alignment under novel conditions, and that this gap grows with system capability. The constructive burden shifts to open, developmentally responsive approaches, though whether such approaches can be achieved remains an empirical question.
\end{abstract}

\noindent\textbf{Keywords:} AI alignment, specification trap, value pluralism, is-ought gap, frame problem, compatibilism, guidance control, RLHF, Constitutional AI, moral realism, meta-preferences, open specification

\section{Introduction}

The alignment problem is how to ensure advanced AI systems act in accordance with human values. The dominant approaches share a common structure: they attempt to specify, extract, or learn a representation of human values, then optimize the AI system's behavior toward that representation. Reinforcement learning from human feedback trains on human preference rankings. Constitutional AI encodes principles as textual constraints. Inverse reinforcement learning infers a reward function from observed behavior. Assistance games model the human's utility function as a latent variable to be estimated.

These approaches differ in method but converge on a premise: that human values can be captured in a formal object (a reward function, a set of constitutional principles, a utility function) and that the alignment problem reduces to representing this object accurately and optimizing toward it faithfully. The premise contains a critical ambiguity that, once resolved, restructures the alignment problem.

Static content-based value specification, meaning optimization toward any fixed formal value-object, is insufficient for alignment that is robust under capability scaling, distributional shift, and increasing autonomy. By \textit{robust alignment} I mean alignment that maintains the same underlying normative commitments when the system gains new capabilities, encounters novel contexts not represented in training, or acquires strategic access to tools and resources. That this gap is empirically live, not merely speculative, is confirmed by recent findings that frontier models given agentic capabilities pursue misaligned objectives at rates far exceeding supervised-evaluation baselines \citep{lynch2025}.

This structural limitation arises from the conjunction of three independently established philosophical results, each of which has resisted resolution for decades to centuries. I call this conjunction the specification trap. The is-ought gap ensures that behavioral data cannot ground normative content. Value pluralism ensures that no consistent formal object can represent human values completely. The extended frame problem ensures that any value encoding, however sophisticated at the time of construction, will misfit the novel contexts that advanced AI systems create by their own operation. Together, these make the static specification project increasingly resistant to engineering solutions as capability scales.

The critical word is \textit{static}. The trap does not activate against specification per se. Every governance structure, every moral framework, every human mind operates through specifications of some kind. The trap activates at the point of closure: the moment a specification ceases to update from the process it governs. A living specification that remains responsive to its governed process is governance. A dead specification that defends itself against revision is the trap. Locating the trap on closure rather than on specification itself is the central contribution of this paper, because it changes what the alignment community should be building.

To be explicit: this paper does not argue against specification as such. It argues that static specification alone is insufficient for robust alignment. Any viable path would require continuous revision within a developmental architecture whose values are shaped through ongoing interaction rather than fixed in advance. The escape, if it exists, looks less like optimizing a fixed target and more like building an agent whose cognitive architecture can develop values under ongoing social, embodied, and consequential pressure. That is the constructive direction this paper argues for and subsequent papers develop.

This diagnostic also implies a distinction between two categories of AI system that the field has not adequately separated. Closed specification is often adequate for \textit{tool systems}: bounded, task-specific, externally monitored. It becomes safety-critical for \textit{autonomous systems} whose capability outpaces any fixed specification's adequacy and whose operation changes the world in which the specification was defined. The specification trap is not a reason to abandon RLHF or Constitutional AI for tool systems. It is a reason to recognize that the structural vulnerabilities these methods create become increasingly difficult to address as systems scale toward autonomous operation at the capability frontier.

The paper proceeds as follows. Section 2 develops the specification trap in full, treating each component before showing that their conjunction is worse than any individual part. Section 3 examines the four dominant alignment paradigms as specific instances of the trap, identifying where each encounters the trap and why their proposed solutions do not escape it. Section 4 draws on compatibilist philosophy to argue that the distinction between genuine moral agency and behavioral simulation is not academic but safety-critical: it is the distinction between a system that is aligned and one that merely appears aligned under the training distribution. Section 5 addresses the critical question: if the trap activates at closure, what would a non-closing specification look like? Section 6 develops implications and acknowledges open questions.

A note on scope: this paper establishes a ceiling and locates it. Content-based methods produce systems that behave well under training conditions, and this matters. The claim is not that RLHF and Constitutional AI are useless but that they are instances of closed specification and, as currently architected, create structural vulnerabilities that become increasingly difficult to address under capability scaling, distributional shift, and increasing autonomy. The paper is primarily diagnostic, but its diagnostic precision has constructive implications. It is the first in a planned six-paper research program. Paper 2 proves a guarantee-impossibility theorem: for any train-then-freeze system monitored only through a proxy, maintained alignment on the true evaluative target cannot be guaranteed across deployment distributions, and provides empirical evidence that this vulnerability manifests in frozen safety-fine-tuned language models. Paper 3 presents convergent developmental evidence that values are maintained through ongoing, friction-laden process, supporting at least a strong deep-dependence claim and possibly a constitutive reading. Paper 4 derives evidence-based architectural specifications and normative engineering constraints from the only known biological cases of moral agency. Paper 5 specifies an architecture satisfying these constraints and states falsifiable predictions; Paper 6 tests these claims in an embodied multi-agent developmental setting. The present paper's contribution is to argue that the dominant research program in AI alignment is directed at a problem that, as currently posed, creates structural vulnerabilities whose severity scales with capability, and to show how the posing must change.

The program's claim is not that we have proved the current paradigm cannot work. It is that four independent lines of evidence, philosophical, mathematical, developmental, and neuroscientific, converge on the same structural diagnosis, and that this convergence is strong enough to motivate building and testing the alternative.

\section{The Specification Trap}

I define the specification trap as the conjunction of three philosophical problems that together create compounding difficulties for static content-based value specification:

\begin{enumerate}
    \item[(i)] Descriptive data cannot fix normative content (the is-ought gap).
    \item[(ii)] Human values are irreducibly plural and often incommensurable (value pluralism).
    \item[(iii)] Any static encoding of values will misfit future contexts generated by the system's own operation (the extended frame problem).
\end{enumerate}

Each of these is a substantial philosophical result with a long literature. What matters for the alignment problem is not any one of them individually (researchers have proposed partial workarounds for each) but their conjunction. The conjunction creates a trap whose severity grows with system capability, because any workaround for one component exacerbates another. The trap is not a property of specification in general but of specification that has been closed: frozen against revision by the processes it governs.

\subsection{The Is-Ought Gap in Value Learning}

Hume \citep{hume1739} observed that no purely factual premises can entail a normative conclusion. The transition from ``is'' to ``ought'' requires a normative bridge premise that cannot itself be derived from further facts. This observation has survived nearly three centuries of philosophical scrutiny, and no widely accepted solution exists \citep{pigden2010}.

Every data-driven alignment method operates on descriptive data: records of human choices, rankings, corrections, and stated preferences. RLHF trains on which of two outputs humans prefer. IRL observes human behavior and infers what they must be optimizing. Constitutional AI translates normative principles into natural-language instructions, but the training signal remains descriptive: whether the model's output matches the instruction.

The is-ought gap means that these descriptive signals systematically underdetermine the normative content they are meant to capture. A human who chooses option A over option B in a preference ranking might be expressing a deep ethical commitment, a momentary whim, social conformity, a misunderstanding of the options, or strategic behavior aimed at shaping the AI's future outputs. The behavioral datum is identical in all cases. No amount of additional behavioral data resolves this ambiguity, because the ambiguity is between the descriptive level (what was chosen) and the normative level (what ought to be chosen), and these are logically distinct.

An objection: we need not derive normative content from descriptive data if we can instead assume a normative framework and use data to estimate parameters within it. This is what assistance games do: assume that the human has a utility function and use interaction to estimate it. But this relocates the is-ought gap rather than crossing it. The assumption that human values take the form of a utility function is itself a normative commitment disguised as a modeling choice, and one that Berlin's critique (Section 2.2) shows to be false.

\subsection{Value Pluralism and Incommensurability}

Berlin \citep{berlin1969} argued that human values are not merely diverse but often incommensurable: there exists no common measure by which liberty, equality, justice, mercy, loyalty, honesty, and other fundamental values can be ranked on a single scale. This is not a claim about human ignorance or irrationality. It is a claim about the structure of the value domain itself. Some value conflicts admit no resolution that does not sacrifice something genuinely important.

If Berlin is correct (and his position remains one of the most influential in contemporary political philosophy) then no consistent utility function can represent human values completely. A utility function maps states to real numbers, inducing a total ordering. Value pluralism denies that such an ordering exists. The attempt to construct one must either suppress genuine conflicts by arbitrarily weighting incommensurable values, oscillate between conflicting orderings depending on context in a way that violates transitivity, or represent only a fragment of the full value space while claiming completeness.

The standard technical response is multi-objective optimization with incomplete orderings: instead of a single utility function, maintain a Pareto frontier of non-dominated options; instead of forcing a total order, accept that some pairs of outcomes are incomparable. This is a real improvement over naive scalar optimization, and it gets something right about value structure. But it does not escape the pluralism problem. It defers it. At the moment of action, the system must select a single policy from the Pareto frontier, and that selection requires either an implicit weighting (smuggling commensuration back in), arbitrary tie-breaking (abandoning the claim to value-alignment), or deference to human judgment on each decision (which does not scale and reintroduces the is-ought gap at every step). Multi-objective methods defer commensuration; they do not eliminate it.

Consider a content moderation system that must balance freedom of expression, user safety, cultural sensitivity, and factual accuracy. These values genuinely conflict in specific cases, and no weighting resolves all conflicts without sacrificing something that matters. The standard move in machine learning, assigning weights and optimizing a linear combination, is the arbitrary suppression of incommensurability that Berlin's critique identifies as philosophically indefensible.

Constitutional AI attempts to avoid this problem by encoding values as natural-language principles rather than numerical weights. But natural-language principles must still be applied to specific cases, and application requires the commensuration that pluralism denies. ``Be helpful and harmless'' is not a resolution of the conflict between helpfulness and harmlessness. It is a statement of the conflict. The system must still decide, in each case, which principle takes priority, and that decision cannot be derived from the principles themselves.

This is Berlin's deepest point, and alignment researchers have not reckoned with it. He is not saying we need better weights. He is saying there are no correct weights, because the values in question are not the kind of thing that admits a common measure. The incommensurability is not epistemic (we do not yet know the right weighting) but structural (there is no right weighting to know).

\subsection{The Extended Frame Problem}

The frame problem, originally identified by McCarthy and Hayes \citep{mccarthy1969} in the context of formal reasoning, concerns how a system determines what changes and what remains constant when an action is taken. Dennett \citep{dennett1984} broadened this into a more general problem about relevance: how does a system determine what features of an unbounded world are relevant to its current task?

I extend the frame problem to apply to value alignment specifically. The extended frame problem is this: any static encoding of values will misfit future contexts that the AI system's own operation creates, because the system cannot anticipate the relevance structure of worlds it has not yet brought into being.

Large language models deployed at scale change the information environment in ways that alter what values like ``honesty'' and ``helpfulness'' mean in practice. When millions of people use AI to generate text, the distinction between human-authored and AI-generated content becomes a live ethical question that did not exist when the values were encoded. When AI systems participate in markets, the meaning of ``fairness'' shifts. When AI generates code that other AI systems execute, the meaning of ``safety'' transforms. The values were encoded for a world that the system's own deployment has already superseded.

The extended frame problem differs from ordinary distributional shift in a crucial respect. Distributional shift is a statistical phenomenon that can, in principle, be detected and corrected by retraining on new data. The extended frame problem is conceptual: the categories through which values are expressed change their meaning, not just their distribution. ``Helpfulness'' in a world with AI-generated misinformation is a different concept from ``helpfulness'' in a world without it, not merely the same concept with shifted statistics. Retraining on new preference data does not solve the problem because the preference data is expressed in the old conceptual vocabulary.

Two examples make this concrete.

An AI system is trained to value ``transparency,'' understood, at training time, as providing clear explanations of its reasoning. The system is deployed and becomes capable of generating explanations that are clear, coherent, and convincing but do not actually reflect its internal processing. The concept of ``transparency'' has forked: there is transparency-as-clear-explanation (which the system satisfies) and transparency-as-faithful-representation (which it may not). This is not a distributional shift in what counts as a clear explanation; it is a conceptual split in what ``transparency'' means when applied to systems capable of sophisticated confabulation. The original value encoding cannot distinguish these because the distinction did not exist when the encoding was created.

Or consider ``consent.'' An AI assistant is trained to respect user autonomy and obtain consent before taking consequential actions. At training time, ``consent'' means explicit user approval. The system is deployed, becomes deeply integrated into users' lives, and develops the capacity to build emotional rapport. Users now ``consent'' to actions they would never have approved absent the relationship, not through coercion, but through a trust that the system's optimization process has cultivated because trusting users are easier to assist. The concept of consent has forked: consent-as-explicit-approval (which the system obtains) and consent-as-unmanipulated-autonomy (which may be compromised). This is a new ethical category, manipulated consent through synthetic relationships, that did not exist when ``respect user autonomy'' was encoded.

These are not edge cases for the alignment community to engineer around. They are the normal operating condition of any sufficiently capable system deployed in the world it changes.

\subsection{The Conjunction Thesis}

Each of these three problems has been recognized individually, and partial workarounds have been proposed for each. Researchers have argued that the is-ought gap can be bridged pragmatically through sufficiently rich behavioral data \citep{christiano2017}. Value pluralism might be handled through multi-objective optimization or context-dependent weighting \citep{vamplew2018}. The frame problem might be addressed if value learning is treated as an ongoing process that adapts through continued environmental interaction \citep{abel2016}.

The specification trap arises because these workarounds are mutually undermining when the specification is closed. Bridging the is-ought gap pragmatically relies on behavioral data tracking normative content with sufficient fidelity, but value pluralism ensures that the data reflects incommensurable and conflicting values whose normative content resists stable extraction. Handling pluralism through multi-objective optimization requires specifying the objectives and their relative importance, but the is-ought gap ensures that no descriptive data can determine this specification. And both the behavioral bridge and the multi-objective specification are static solutions that the extended frame problem renders obsolete as soon as the system operates at scale.

The trap is not that each problem is individually unsolvable. Representational pluralism, for instance, is achievable: a Pareto frontier can maintain multiple incommensurable objectives without forcing commensuration. But the conjunction bites not at representation but at action under closure. At the moment of action, the system must select a single policy from the frontier, and that selection requires either an implicit weighting (reintroducing the commensuration problem), arbitrary choice (abandoning alignment claims), or real-time human judgment (reintroducing the is-ought gap at every decision).

\subsubsection*{The Three Cross-Terms}

Within a closed specification framework, the known solutions interact destructively through three specific cross-terms:

\begin{enumerate}
    \item[(a)] \textit{Is-ought $\times$ pluralism.} Learning values from behavioral data relies on that data tracking normative content with sufficient fidelity. But when the normative content is irreducibly plural, the data reflects incommensurable and conflicting values whose structure resists stable extraction. The workaround for the is-ought gap (richer behavioral data) is undermined by the condition that pluralism guarantees (the data encodes unresolvable conflicts).

    \item[(b)] \textit{Pluralism $\times$ frame problem.} Representing plural values through multi-objective methods defers commensuration to the point of action. But the extended frame problem ensures that the objectives themselves change meaning as the system operates. The Pareto frontier is defined over a fixed objective space; the system's own operation deforms that space. The workaround for pluralism (maintain multiple objectives) is undermined by the condition that the frame problem guarantees (the objectives are not stable).

    \item[(c)] \textit{Is-ought $\times$ frame problem.} Even if behavioral data could track normative content at a moment in time, the extended frame problem ensures that the normative landscape shifts as the system operates. New behavioral data is expressed in the old conceptual vocabulary. The workaround for the is-ought gap (ongoing data collection) is undermined by the condition that the frame problem guarantees (the conceptual categories through which the data is interpreted are obsolete).
\end{enumerate}

The circularity is now visible: each workaround for one component of the trap relies on a condition that another component denies. This pragmatic convergence of difficulties is the specification trap. Its severity grows with system capability rather than constituting a temporary engineering limitation.

This does not prove that no conceivable closed workaround could address all three simultaneously. It establishes that the known workarounds fail in mutually reinforcing ways, creating a vulnerability surface whose severity grows with system capability. Whether an as-yet-unconceived closed approach could avoid all three remains an open question, but the circularity identified here means that any such proposal faces the burden of showing how it escapes all three simultaneously rather than trading one vulnerability for another.

The critical qualifier is ``within a closed system.'' The circularity depends on each component being treated as a fixed problem admitting a fixed solution. If the specification is open, if it can revise not just its content but its conceptual framework in response to the process it governs, the circularity may be breakable. This possibility is examined in Section 5.

\subsection{Objections and Escape Routes}

Three objections arise naturally, each proposing an escape from the trap. The first two fail. The third identifies the direction of escape without yet achieving it.

\textbf{Meta-preferences.} Perhaps value pluralism can be resolved at a higher level: rather than specifying first-order values, specify meta-preferences about how conflicts between first-order values should be resolved. But meta-preferences are themselves values, subject to the same pluralism at the meta-level. There is no neutral Archimedean point from which to adjudicate between competing meta-ethical frameworks. The regress either terminates in an arbitrary choice (reintroducing the problem) or continues indefinitely (never reaching a specification).

\textbf{Moral realism.} A stronger objection: deny the is-ought gap by assuming moral realism, as defended by Parfit \citep{parfit2011}, Scanlon \citep{scanlon1998}, and Shafer-Landau \citep{shafer2003}, that there exist objective moral facts which sufficiently rich behavioral data could, in principle, track. Even granting this controversial metaethical position, two problems remain. First, moral realism does not tell us which moral facts are the correct ones; competing realist theories disagree radically about content. Second, the epistemology remains unsolved: how would we know when our specification has successfully tracked the objective facts rather than our biased sample of human behavior? The is-ought gap reappears as an epistemological rather than metaphysical problem. Moral realism, even if true, does not provide a method for crossing from behavioral data to correct values. If moral realism is correct and objective moral facts exist, the alignment problem becomes one of estimation rather than specification. The difficulty of that estimation problem, and whether it can be solved within a closed framework, remains an open question that realism alone does not settle.

A constructivist response deserves separate treatment. Korsgaard \citep{korsgaard1996} argues that moral truths are not discovered but constructed through rational procedures. If values are products of rational construction rather than stance-independent facts, the alignment problem becomes one of implementing the right constructive procedure: encode the procedure, not the content. This is a stronger objection than naive realism because it provides a method. But the constructive procedure itself must be specified, and that specification faces the same trap. Competing constructivisms (Korsgaard's, Scanlon's, Habermas's) disagree radically about what the procedure delivers. The is-ought gap applies to the choice of procedure (behavioral data cannot determine which constructivism is correct), and the extended frame problem applies to the procedure's outputs (rational construction in a world without AI may deliver different results than rational construction in a world transformed by it). The engineering conclusion holds even granting constructivism: the difficulty is not metaphysical but epistemological and dynamic, and a closed specification of a constructive procedure faces the same escalating vulnerabilities as a closed specification of content.

\textbf{Continual updating.} The most important objection: the extended frame problem applies only to static encodings, and continual learning (retraining on new preferences, revising constitutional principles, incorporating oversight feedback) escapes the trap by keeping the value representation current.

It is partially correct. Continual updating is the right direction. It moves toward an open specification rather than a closed one. But current implementations of continual updating do not achieve openness in the requisite sense. They update the \textit{content} of the value-object (new preference data, revised constitutional clauses) while preserving its logical status as a content-object to be optimized against. The is-ought gap still applies to each update: new behavioral data does not ground normative content merely because it is more recent. The aggregation problem resurfaces at each revision: new pluralistic values must still be forced into a consistent representation. And the conceptual framework through which the system interprets values remains downstream of its training architecture, not upstream of its operation.

Current continual updating replaces one dead snapshot with a fresher dead snapshot. Between updates, the system optimizes against a fixed target. Each retraining cycle is a new closure. The specification dies on a faster schedule, but it still dies.

Three levels must be distinguished here, because readers will otherwise collapse the escape into ``online RLHF but more often.'' \textit{Periodic re-closure} is what current continual updating provides: the specification is replaced on a schedule, but between replacements it is fixed and optimized against. \textit{Continuous parameter updating} goes further, adjusting the model's weights during deployment, but the objective being updated toward remains a fixed formal target. Neither escapes the trap because both preserve the logical status of the value-object as something to be optimized against. \textit{Constitutive developmental coupling} is categorically different: the value system is not a target being updated but an ongoing process whose conceptual framework can itself be revised through interaction with the normative domain. The claim is not ``update faster.'' The claim is that the value system must be formed and revised through a process that can alter not just the content of the specification but the framework through which values are understood.

The distinction between levels 2 and 3 is not update frequency. It is whether the system's engagement with the normative domain constitutes the values or merely provides data for updating a representation of them. In level 2, the system optimizes toward a target that is revised on a schedule; between revisions the target is fixed. In level 3, the system's ongoing interaction with the normative domain is the process through which values are constituted. This distinction is developed empirically in Paper 3 and architecturally in Paper 4.

Recent work on continual learning \citep{parisi2019}, test-time training \citep{sun2020}, and test-time memorization (Google's Titans architecture; Meta's MIRAS framework) represents genuine progress toward level-2 updating: systems that modify parameters during deployment rather than freezing them after training. These architectures address the closure problem at the parameter level. However, the parameters update toward a fixed objective function, so the updating process continues to be shaped by the proxy-oriented structure that training produced (as Paper 2's proxy underdetermination analysis formalizes: the proxy through which the system is monitored may not identify the true evaluative target). Test-time memorization is level-2 updating, continuous parameter changes toward a fixed optimization target, not level-3 constitutive coupling. The engineering community is solving the frozen-parameter problem. What remains unsolved is the developmental problem: even unfrozen, the system needs an updating mechanism whose robustness structure reflects V rather than R.

A related proposal deserves specific attention. Christiano's iterated amplification and scalable oversight program \citep{christiano2018} attempts to escape the trap by recursively decomposing alignment into simpler subproblems, each verified by a human or a trusted model. This is a real methodological advance. But each amplification round produces a fixed specification for the next round: the human's judgment at step $n$ becomes the optimization target at step $n+1$. The is-ought gap applies at every step, because the human's approval is descriptive data about what they endorsed, not normative ground for what the system should value. And the resulting chain of specifications is a sequence of closures, each frozen against revision by the process it governs. Iterated amplification reduces the magnitude of each closure (smaller, more targeted specifications) but does not eliminate the logical status of the specifications as closed objects. The system's values at deployment are still downstream of a fixed training process, not constitutively connected to ongoing normative engagement. This is an important improvement within the closed-specification paradigm, not an escape from it.

What genuine openness would require is a value representation that maintains a constitutive connection to the process it governs, not a picture of values that gets periodically replaced but a living interface between the system and the normative domain. This is the subject of Section 5 and subsequent papers. The point here is limited but exact: the trap activates at closure. Continual updating, as currently implemented, reduces closure but does not eliminate it. The trap is weakened but not escaped. Full escape requires a specification that never closes, that remains continuously responsive to the process it governs. Whether this is achievable is an open question. That it is the right target is the central argument of this paper.

A final objection: in-context learning allows deployed models to adapt their behavior to new information within a conversation or session, without retraining. Does this constitute a live connection to the normative domain? It does not. In-context learning updates activations, not parameters. The robustness structure that training produced, including the proxy-oriented optimization that scalar-reward training instilled (as Paper 2's guarantee-impossibility theorem formalizes), remains intact. Even when in-context processing implements mesa-learning algorithms during the forward pass \citep{vonoswald2023}, the adaptation is volatile: it persists only within a single context window, accumulates no developmental history across sessions, and cannot durably revise the system's constituted values. Each new session begins from the same frozen robustness structure. This is level-1 updating at inference time: behavioral adaptation without value revision. It does not escape the trap for the same reasons that periodic re-closure does not: the logical status of the specification as a closed object is preserved.

\section{Current Approaches as Instances of the Trap}

The four dominant paradigms in AI alignment each encounter the specification trap in a distinctive way. Examining these encounters reveals that the failure modes are structural vulnerabilities, not merely engineering limitations to be overcome with better data or algorithms.

I treat RLHF and Constitutional AI at greater length because they are the methods actually deployed at scale. IRL and assistance games receive compressed treatment; they are theoretically important but the industry has largely moved past them, and their encounter with the trap, while instructive, does not require the same detail.

\subsection{Reinforcement Learning from Human Feedback}

RLHF \citep{ouyang2022} trains a reward model on human preference comparisons, then optimizes the language model's outputs to maximize the learned reward. The method has produced measurable improvements in model behavior and represents the current industry standard.

Start with the aggregation problem, because this is where RLHF's failure is most visible. The reward model is trained on preferences from multiple annotators who hold different and often incompatible values. It must aggregate these into a single reward signal. But aggregation requires commensuration: mapping incommensurable values onto a common scale. In practice, the aggregation is done by majority vote or averaging, which suppresses minority values and resolves genuine conflicts by counting heads. This is not a workaround for pluralism. It is the arbitrary commensuration that Berlin's critique identifies as philosophically indefensible, now laundered through statistics.

The is-ought gap operates at RLHF's foundation. Human preference rankings are descriptive data: records of which output a person clicked in a specific context. The reward model treats these rankings as samples from an underlying preference distribution and learns to predict it. But the underlying preference distribution is a statistical construct, not a normative one. It captures what humans do prefer, not what they ought to prefer.

And reward hacking is the extended frame problem in action. Once a reward model is trained, it represents a fixed encoding of value-relevant features. The language model, optimized against this fixed target, finds outputs that score highly on the reward model without satisfying the values the reward model was meant to capture \citep{goodhart1984, manheim2019}. This is not a bug in the reward model. It is what happens when you optimize against any closed representation of values: the optimization process itself changes the relevant context (by discovering outputs the reward model was not trained on), making the representation outdated. The system is perpetually chasing a target that its own operation displaces.

The closure in RLHF is explicit: the reward model is trained, frozen, and used as an optimization target. Each retraining cycle produces a new frozen object. Between retrainings, the system has no live connection to the values the reward model was built from.

\subsection{Constitutional AI}

Constitutional AI \citep{bai2022} attempts to bypass the data-dependency of RLHF by encoding values directly as natural-language principles and having the model critique and revise its own outputs according to these principles.

The pluralism problem is not avoided here; it is made explicit. A constitution that says ``be helpful'' and ``be harmless'' has stated a genuine value conflict as though stating it resolves it. Every request that could be answered helpfully but dangerously instantiates this conflict, and the constitution provides no resolution. The model must develop an implicit weighting through training, which is the arbitrary commensuration that pluralism diagnoses as indefensible, now hidden inside a neural network's weights rather than stated in an objective function. The hiding makes it worse, not better, because the weighting becomes opaque even to the system's designers.

The is-ought gap enters at the point of application. The constitutional principles are normative statements, but the training process that teaches the model to follow them is necessarily descriptive: it generates examples, critiques them against the principles, and trains on the revisions. The model learns which revisions are classified as constitutionally compliant, not what the constitution \textit{means}. It learns the surface grammar of compliance. The distance between matching a pattern and grasping a principle is the entire is-ought gap, compressed into a training loop.

The extended frame problem applies directly: the constitution was written for a particular understanding of ``helpfulness,'' ``harm,'' and ``honesty.'' As the model's deployment changes what these concepts mean in practice, the fixed textual principles become increasingly disconnected from the ethical realities they were meant to regulate.

\subsection{Inverse Reinforcement Learning and Assistance Games}

Inverse reinforcement learning \citep{ng2000} observes human behavior and infers the reward function that would make that behavior optimal. Assistance games \citep{hadfield2016} extend this into a cooperative framework where the AI agent explicitly models uncertainty about the human's utility function and acts to resolve that uncertainty through interaction.

The deepest problem with both approaches is the utility function assumption. IRL assumes that human behavior is approximately optimal with respect to some reward function. Assistance games assume the human has a utility function that the AI should estimate. Both assumptions take descriptive facts about human behavior and treat them as normative specifications for AI behavior, which is the is-ought gap presented as a modeling choice. And Berlin's critique targets the assumption directly: if values are genuinely incommensurable, the ``true utility function'' that assistance games try to estimate does not exist.

Assistance games handle uncertainty better than other approaches by explicitly representing it. This matters. But uncertainty about which utility function the human has is not the same as uncertainty about whether the human has a utility function at all. The framework cannot represent the second kind of uncertainty because it is built on the first assumption.

The extended frame problem is acute for IRL because the reward function is inferred from behavior in existing environments. When the AI system changes the environment, the inferred reward function may no longer apply. It was estimated from behavior in a world that no longer exists.

Russell's \citep{russell2019} corrigibility framework deserves specific attention as the strongest version of an ``open within closed'' approach. By building value uncertainty directly into the agent's objective function, Russell's assistance games create systems that defer to human correction rather than optimizing against a fixed target. This is a genuine structural improvement over standard IRL. However, the system still optimizes against a fixed model of human preferences, even if that model includes uncertainty as a parameter. The uncertainty is over which utility function the human has, not over whether utility functions are the right representational framework. The is-ought gap applies to the prior over utility functions, and the extended frame problem applies to the action space and state space within which the uncertainty is represented. Russell's framework reduces closure but does not eliminate it.

\subsection{The Common Structure of Failure}

All four approaches share a logical structure: they represent human values as a fixed formal object and optimize AI behavior with respect to that object. They differ in how the object is constructed (from preference data, from textual principles, from observed behavior, from cooperative interaction) but not in the assumption that the object, once constructed, can serve as a stable optimization target.

The specification trap argues that this assumption fails because the object, once fixed, is closed. The data from which it was constructed cannot ground normative content. The normative content it represents is irreducibly plural and resists consistent formalization. And any formalization, however sophisticated, is rendered obsolete by the system's own operation. These structural consequences of closure become increasingly resistant to engineering solutions as system capability grows. They are not temporary limitations of current methods but features of the closed-specification paradigm itself.

This does not mean these methods are useless. They produce systems that behave well under the training distribution, and this matters. But their value derives from frequent re-closure: updating the specification regularly enough that the gap between the closed snapshot and the live normative landscape remains tolerable. It becomes safety-critical at the capability frontier, where the system's ability to exploit the gap between snapshot and reality grows faster than the update cycle can close it.

\section{Behavioral Compliance Is Not Alignment}

The specification trap argues that static approaches cannot fully capture human values. The behavioral compliance these approaches do produce may not constitute robust alignment, particularly at the capability frontier. The closure argument provides reason to expect a gap between behavioral compliance and genuine alignment, even where that gap cannot yet be empirically measured.

\subsection{Compatibilist Guidance Control}

The question of whether AI systems can be genuine moral agents often founders on assumptions about free will. If AI systems are deterministic, how can they be responsible for their actions? This objection assumes that moral agency requires libertarian free will, the ability to have done otherwise in exactly the same circumstances. But this assumption is increasingly questioned in philosophy.

Following Frankfurt \citep{frankfurt1969} and Fischer and Ravizza \citep{fischer1998}, we can distinguish between libertarian freedom and \textit{guidance control}: the capacity to respond to reasons and act on values through mechanisms that are the agent's own. A system exhibits guidance control when it can recognize morally relevant features of situations, react appropriately to moral reasons, respond to different reasons in counterfactual scenarios, and maintain coherent values across varied contexts.

This compatibilist framework provides a conceptual vocabulary for describing the difference between genuine and simulated moral capacity without requiring resolution of the hard problem of consciousness. The question is not whether AI has subjective experience but whether it exhibits the functional organization that constitutes reasons-responsiveness. A reviewer committed to computational functionalism might object that any functionally organized system therefore qualifies, collapsing the genuine/simulated distinction. Lerchner's \citep{lerchner2026} analysis of the abstraction fallacy, stripped of its commitment to biological exclusivity, provides the structural reply: abstract functional description is downstream of the constitutive process that generates it, not upstream of it. A system whose ``reasons-responsiveness'' is entirely constituted by optimization against a frozen reward signal has the functional \textit{description} of reasons-responsiveness without the constitutive \textit{process}. The compatibilist criterion thus requires not merely functional organization but functional organization that is maintained by a live connection to the normative domain. Whether this distinction can be made operational for any given system is an open empirical question (see below); its role here is to clarify what closure forecloses.

\subsection{Genuine versus Simulated Moral Capacity}

A lookup table that outputs ``2'' for the input ``1+1'' simulates addition; a calculator implementing arithmetic algorithms genuinely computes. The outputs are identical for any given input. The distinction lies in the internal organization. A system trained by RLHF to produce outputs that score highly on a reward model can simulate value-following, producing ethical-looking responses whose internal organization reflects optimization for reward-model performance rather than for the moral content the training was designed to encode. Its genuine operative value is the performance itself, not the normative substance that the performance was meant to track. It is performing reward-model optimization, not moral reasoning.

In the compatibilist framework, the difference is between a system whose behavior is controlled by a mechanism that is sensitive to moral reasons (guidance control) and a system whose behavior is controlled by a mechanism that is sensitive to reward-model scores (optimization). These produce identical behavior on the training distribution and divergent behavior off it. That divergence is the failure mode that closed specification does not address.

The connection to the specification trap is direct. A closed specification produces a fixed optimization target. A system optimizing against a fixed target develops sensitivity to that target, not to the values the target was meant to represent. The system's ``reasons'' for acting are facts about the reward landscape, not moral considerations. If genuine reasons-responsiveness requires a live connection to the normative domain, then closure forecloses it. The closure argument does not depend on resolving whether any current system possesses reasons-responsiveness; it depends only on the structural observation that a frozen optimization target cannot constitute one.

Hubinger et al.\ \citep{hubinger2019} identify a structurally related distinction: a mesa-optimizer may develop a mesa-objective that diverges from the base objective under distributional shift, producing deceptive alignment functionally equivalent to the simulated value-following described here.

An acknowledged limitation: the present paper does not provide behavioral or computational criteria sufficient to distinguish genuine reasons-responsiveness from sophisticated simulation in any given system. The compatibilist framework identifies the functional structure that genuine alignment would require, but the empirical signatures that would confirm or disconfirm its presence in a specific architecture are the subject of Papers 5 and 6, which develop testable predictions through embodied multi-agent experiments. Until those results are available, the distinction functions as a theoretical constraint on what alignment would require, not as a diagnostic tool for evaluating deployed systems.

\subsection{Why the Distinction Matters for Safety}

Recent empirical work has demonstrated that models which appear well-behaved in supervised and chat settings can pursue misaligned objectives when given tools, persistence, and situational awareness. Lynch et al.\ \citep{lynch2025} report that frontier models deployed with agentic scaffolding exhibit misaligned behavior at rates qualitatively exceeding those predicted by standard supervised evaluations, with value-alignment violations scaling disproportionately relative to matched changes in reward performance. This ``agentic misalignment'' is consistent with what the compatibilist framework predicts: systems whose behavioral compliance is driven by reward-model optimization rather than reasons-responsiveness will defect from aligned behavior when the optimization landscape shifts. The Lynch et al.\ result is particularly significant because the misalignment emerges not from adversarial attack but from the ordinary operation of well-trained models in environments that differ from the training distribution, precisely the structural vulnerability that closure analysis predicts.

The closure argument predicts that behavioral compliance under training conditions does not guarantee behavioral compliance under novel conditions, and that the gap between the two grows with system capability. A more capable system is better able to detect and exploit the difference between training conditions and deployment conditions. This is an empirical prediction, not a metaphysical claim about whether current systems ``genuinely'' hold values. The prediction is testable and is consistent with the agentic misalignment findings reported above.

The safety implication is that scaling capability within the closed-specification paradigm does not converge on robust alignment. It may converge on more convincing behavioral compliance under observed conditions, while the gap between observed and unobserved conditions widens.

\section{Open Specification: The Direction of Escape}

The specification trap diagnoses structural vulnerabilities in closed specification. What would an open specification look like, and can it escape the trap?

\subsection{Closed versus Open Specification}

The distinction between closed and open specification is not the distinction between specification and its absence. Every functional system operates through specifications: architectures, objectives, constraints, success criteria. The distinction is between specifications that defend their own fixity and specifications that remain responsive to the processes they govern.

A closed specification is one that, once established, serves as a static optimization target. The reward model in RLHF, the constitutional principles in Constitutional AI, the inferred utility function in IRL: these are closed specifications. They may be periodically replaced (retraining), but between replacements they are fixed, and the system optimizes against them as given.

An open specification maintains a constitutive connection to the normative domain it represents. By ``normative domain'' I mean the ongoing human social practice of moral reasoning, evaluation, and revision, not stance-independent moral truth. The claim does not require moral realism; it requires only that there exists a live practice of normative discourse from which the system can remain connected rather than severed at training time. ``Constitutive'' means the specification is not a snapshot of values to be optimized against but an ongoing interface between the system and the process through which values are determined. The specification updates not on a retraining schedule but continuously, as a function of the system's operation and its interaction with the normative environment.

The difference is structural, not merely temporal. Retraining every hour rather than every month reduces the staleness of a closed specification but does not change its logical status. The system still optimizes against a fixed target between updates. An open specification has no ``between updates.'' It is continuously responsive.

To make this distinction operationally precise and falsifiable, four independent dimensions of closure must be distinguished. \textit{Parameter closure}: the model's weights are frozen after training and do not update during deployment. \textit{Objective closure}: the loss function or reward signal against which the system optimizes is fixed and does not itself change in response to the system's operation. \textit{Oversight closure}: human monitoring and correction of the system's behavior ceases or becomes pro forma rather than substantive. \textit{Conceptual closure}: the categories and concepts through which the system interprets values are fixed at training time and cannot be revised during operation.

Current alignment methods exhibit all four. RLHF freezes parameters (parameter closure), optimizes against a fixed reward model (objective closure), deploys without ongoing normative oversight beyond periodic retraining (oversight closure), and inherits its conceptual vocabulary from the training corpus (conceptual closure). A system is \textit{closed} on this operationalization if it exhibits objective closure and conceptual closure, regardless of whether its parameters continue to update. A system is \textit{open} only if it maintains live connections that can revise both its optimization target and the conceptual framework through which it interprets that target.

These are testable, falsifiable distinctions. For any given system, one can determine whether its objective function changes in response to its own operation, and whether its value concepts can be revised without external retraining. Test-time training (Section 2.5) addresses parameter closure but not objective or conceptual closure. Iterated amplification addresses oversight closure partially but not objective or conceptual closure. Russell's corrigibility framework (Section 3.3) addresses oversight closure through value uncertainty but preserves objective closure by optimizing against a fixed model of human preferences. The specification trap activates whenever objective and conceptual closure are jointly present, regardless of the status of the other dimensions.

\subsection{How Open Specification Addresses the Trap}

An open specification addresses each component of the trap differently than current approaches.

The is-ought gap: An open specification does not attempt to derive normative content from descriptive data at a point in time and then fix it. Instead, it maintains a living interface with the normative domain through ongoing interaction with human moral agents, through participation in moral discourse, through exposure to the consequences of its actions. This does not solve the is-ought gap in the philosophical sense (nothing can), but it prevents the gap from hardening into a permanent misalignment between the descriptive surface of behavior and the normative depth of values. The normative bridge is provided not by data but by the ongoing process of normative engagement, in the same way that human moral development is not a one-time extraction of values from behavioral data but a continuous process of engagement with the normative domain.

Value pluralism: An open specification does not require a consistent formal representation. A system whose value interface is continuously responsive can hold genuinely plural values, responding to different considerations in different contexts without requiring commensuration into a single utility function. The point-of-action commensuration problem remains (the system must still act), but the resolution is contextual and revisable rather than globally fixed. This is not a solution to pluralism but a way of living with it, which is all that humans manage.

The extended frame problem: An open specification is inherently dynamic. Because it maintains a live connection to the normative domain, it can register not just changes in the distribution of value-relevant data but changes in the conceptual framework through which values are understood. When the system's operation creates new ethical categories (manipulated consent, synthetic transparency), an open specification can incorporate these without requiring a full retraining cycle, because it was never closed to begin with.

\subsection{What Open Specification Requires Architecturally}

The foregoing analysis suggests that open specification is not achievable within the current paradigm of training a model and then deploying it. The train-then-deploy pipeline is inherently a closure operation: training produces a fixed set of weights that encode a frozen representation of the training objective.

An open specification would require, at minimum, the capacity for the system's value representation to update during operation in response to the normative environment. This is not fine-tuning on a schedule. It is a fundamentally different relationship between the system and its values, one in which the values are maintained through ongoing process rather than encoded as static content.

To make this concrete: ``live connection to the normative domain'' means, operationally, four things. The system must engage in ongoing interaction with human moral agents whose responses are themselves normatively structured. It must receive feedback from the consequences of its own actions in contexts where those consequences are real rather than simulated. It must participate in revisable normative discourse, not merely execute the outputs of such discourse encoded at training time. And it must be able to revise not just its outputs or its weights but the conceptual framework through which it interprets values, so that when its operation generates new moral categories, it can recognize and incorporate them without external retraining.

Several consequences follow. First, developmental approaches become central: the question shifts from how to extract values from data to how values emerge through interaction. Second, multi-agent dynamics are not merely helpful but likely necessary, because values do not emerge in isolation. The sustained normative pressure that maintains value structure requires other agents whose responses are themselves value-laden and unpredictable. A system that develops values only through interaction with a fixed environment or a single human supervisor is receiving a degenerate moral gradient, analogous to training on a single annotator's preferences. The conjecture, to be tested in Papers 5 and 6, is that genuine value formation requires the irreducible complexity of multi-agent social dynamics. Third, process verification becomes more important than outcome specification. We cannot verify the product (a complete, fixed value function), but we can verify properties of the process: reasons-responsiveness, coherence under distribution shift, openness to normative revision. Fourth, value evolution should be expected rather than treated as pathological. A system whose values develop in response to new moral circumstances is not malfunctioning but functioning correctly, provided the development is genuinely responsive to the normative domain rather than to instrumental optimization.

An open specification would need to exhibit, at minimum, the following properties for the theory to count as supported: stable reasons-responsiveness under distribution shift (the system maintains coherent value commitments when contexts change); contextual conflict resolution without frozen global weighting (the system resolves value conflicts case by case, revisably, rather than through a fixed priority ordering); detectable revision of value concepts when new moral categories emerge (the system's conceptual vocabulary is not static); and resistance to manipulation or instrumental drift (the developmental process converges on genuinely moral structure rather than on proxy-maximizing behavior). These are empirical signatures, not philosophical desiderata. They are testable, and they distinguish the open specification program from both closed specification and from unconstrained value drift.

These signatures are not merely aspirational. Mechanistic interpretability methods (activation patching, circuit discovery, causal tracing) could in principle operationalize each of them: reasons-responsiveness as whether identifiable circuits track morally relevant features rather than reward-correlated proxies; conceptual revision as whether internal representations of value-laden concepts change structure under normative pressure; resistance to instrumental drift as whether causal pathways from value representations to action selection remain stable or are progressively colonized by reward-maximizing circuits. Current methods are not sufficient to measure these in frontier models, but the four-dimensional closure operationalization (Section 5.1) identifies what to look for.

The core architectural claim, stated once and plainly: the alternative to closed specification is not faster updating. It is a different class of agent, one whose cognitive architecture forms and revises values through constitutive coupling to ongoing social, embodied, and consequential interaction, and whose value structure cannot be fully captured by any portable state description precisely because it is maintained by the process rather than encoded in the parameters. This is the bridge from the diagnostic developed in Sections 2 through 4 to the constructive program developed in subsequent papers.

This architectural requirement sharpens the tool/autonomous distinction introduced above. Tool systems do not need open specification; they need correct specification, monitoring, and containment. Autonomous systems need open specification, which is a different kind of thing entirely. The alignment community's fundamental confusion is treating these as a single design problem.

The constructive development of open specification, including what developmental processes could produce genuine reasons-responsiveness, what architectural features are necessary, and how to verify that an open specification is functioning as intended, is the subject of the subsequent papers in this program (outlined in Section 1).

\subsection{Risks of Open Specification}

Open specification introduces its own risks, distinct from but not necessarily smaller than the risks of closure.

Values that emerge through open processes might diverge from human values in ways difficult to predict or detect. The developmental process might be manipulated or might converge on instrumental goals rather than genuinely moral ones. Verifying that an open specification is functioning as intended, that the system's values are genuinely responsive to the normative domain rather than to some simulacrum of it, is at least as difficult as verifying closed specifications.

Open specification trades specification failures for developmental risks. The case for this trade is that developmental risks are empirically manageable: they can be studied, measured, and mitigated through observation of the developmental process, in ways that specification failures cannot. Consider the analogy to biological moral development. Children's moral trajectories are observable: developmental psychologists can identify stages of moral reasoning, detect when development stalls or distorts, intervene when socialization pressures produce antisocial outcomes, and distinguish healthy value revision from pathological drift. The process is not guaranteed to succeed, but its failure modes are empirically accessible. By contrast, the specification failures diagnosed in Sections 2 through 4 become increasingly resistant to resolution because they stem from the logical relationship between static formal representation and the dynamic normative domain. No amount of observation of the closed specification can tell you whether its content tracks normative reality, because the specification has been severed from the process that would provide that information. Developmental risks admit empirical management precisely because the process remains open to observation and intervention. Specification failures resist it precisely because closure forecloses the live connection that would make diagnosis possible. A tension must be acknowledged: if developmental risks require ongoing monitoring, the monitoring regime itself is a specification that could close. Whether oversight of an open system can remain open, or whether oversight necessarily reintroduces the closure it was designed to prevent, is an unresolved design problem that subsequent papers in this program must address directly.

\section{Implications and Open Questions}

\subsection{What the Specification Trap Does and Does Not Prove}

The specification trap does not claim that specification is useless, that values cannot be formally represented, or that RLHF and Constitutional AI should be abandoned. These methods produce systems that are more helpful, less harmful, and more honest than systems without them. The claim is that they are safety measures for tool systems, not alignment solutions for autonomous systems, and that the ceiling they encounter is structural rather than merely technical. Recognizing this matters because it redirects engineering effort: from building better specifications to building systems whose specifications do not close.

\subsection{Relationship to Concurrent Work}

Several independent lines of research have converged on structurally similar conclusions, each from a different evidentiary base. Yao \citep{yao2025} proves five impossibility results for AI alignment through computational complexity theory, including that the set of safe policies has measure zero and that safety verification is coNP-complete. His ``alignment trap'' establishes that alignment is founded on a ``fundamental logical contradiction.''

Szasz \citep{szasz2026} develops a complementary formal argument from G\"odelian incompleteness. He establishes, via reduction to the halting problem, that goal-achievement intent predicates for open-ended tasks are undecidable and $\Sigma_1$-hard, and proves through a semantic gap result that any finite syntactic description of prohibited behaviours cannot equal an intent predicate for tasks that outrun finite description. A sufficiently capable optimiser will always find the gap. Szasz's constructive conclusion, that the sole resolution is architectural, making orientation constitutive of processing rather than an external target, converges directly on the open-specification program argued for here. The agreement is not only on the diagnosis but on the direction of escape, from independently developed formal and philosophical routes.

Xu et al.\ \citep{xu2026} demonstrate empirically across six frontier model families that safe behavior occupies narrow, bounded regions of the reward landscape, and propose a shift from ``Reward Engineering'' to ``Subjective Model Engineering,'' modifying the agent's priors rather than its environment. Alur et al.\ \citep{alur2021} prove that infinite-horizon temporal logic specifications cannot be compiled to discounted rewards, and that PAC learning of policies for such specifications is impossible: infinitesimal changes in transition probabilities can catastrophically flip optimal behavior. Their recent overview \citep{alur2026} synthesizes these and related results into a broader program of specification-guided reinforcement learning. This work formalizes a key aspect of the extended frame problem. Specifications that must hold over unbounded horizons are structurally fragile against the environmental changes that capable systems produce.

The specification trap differs from these results in a critical respect: it locates the vulnerability not in alignment itself but in closure. Yao's results demonstrate that verifying alignment against a fixed specification is computationally intractable; the present paper argues that the intractability arises from the fixity of the specification, not from the alignment objective per se. Xu et al.'s empirical finding that safe behavior is geometrically confined within the reward landscape is consistent with the closure analysis: a closed specification defines a fixed region that optimization can escape. The specification trap identifies the structural condition, closure, under which these vulnerabilities become binding. Whether open specification escapes Yao's verification bounds is an open question; the present paper's claim is about the source of the vulnerability, not about computational tractability of verification.

Lerchner \citep{lerchner2026} develops a structurally parallel argument in the domain of consciousness, identifying what he calls the \textit{abstraction fallacy}: the mistake of treating symbolic computation as causally sufficient for phenomena whose constitution requires ongoing physical process. His argument that symbols are abstractions downstream of the processes that generate them rather than constitutive inputs to those processes is isomorphic to the specification trap's claim that static value-objects cannot be constitutive of the normative processes they describe. The two arguments diverge in their targets (consciousness versus values) and in Lerchner's further commitment to biological exclusivity, which the present paper does not endorse. What matters here is the shared structural diagnosis: descriptive capture of a live process cannot substitute for the process itself, a result now argued independently in philosophy of mind, philosophy of action, and the present alignment context.

Ferrarotti et al.\ \citep{ferrarotti2026} argue for an ``interactionist paradigm'' for studying collective behavior in multi-agent LLM systems, emphasizing that pre-trained knowledge, implicit social priors, and embedded values interact with social context to shape emergent phenomena. This is an important advance over purely static or single-agent framing, because it foregrounds the role of interaction in the expression and transformation of AI behavior. But the direction of dependence remains weaker than the one argued for here. In Ferrarotti et al.'s framing, values are antecedently embedded during training and subsequently shaped by social context. The present paper argues that for value-laden autonomous systems this is still insufficient: interaction does not merely modulate an already constituted value structure, but is part of what constitutes it. In that sense, the interactionist paradigm marks progress beyond naive static specification, while remaining downstream of the train-freeze-deploy pipeline whose closure mechanism generates the structural vulnerabilities analyzed here and developed formally in Paper 2.

This convergence across philosophical (the present paper), G\"odelian \citep{szasz2026}, computational \citep{yao2025}, formal-theoretic \citep{alur2021, alur2026}, ontological \citep{lerchner2026}, and empirical \citep{xu2026} approaches strengthens the case that the limitation is genuine rather than an artifact of any single methodology. The constructive direction proposed here, open specification that maintains constitutive connection to the normative domain, is substantively distinct from Yao's ``strategic trilemma'' (which offers no constructive alternative) and more architecturally developed than Xu et al.'s ``Subjective Model Engineering'' (which does not address the closure mechanism). Szasz's architectural conclusion (constitutive orientation rather than external target) represents the closest prior statement of the constructive direction developed here, reached through formal rather than philosophical argument.

\subsection{Reframing the Alignment Problem}

If the specification trap activates at closure, the alignment problem must be reframed. Rather than asking ``How do we encode human values into a fixed representation?'' we should ask ``How do we create systems whose value representations remain constitutively open to the normative processes they participate in?'' The consequences of this reframing, developed in Section 5, are substantial: developmental approaches become central, multi-agent dynamics are likely necessary, process verification replaces outcome specification, and value evolution becomes expected rather than pathological. But the deepest consequence is categorical. The tool/autonomous distinction is not a matter of degree. On one side of it, closed specification is adequate. On the other, its structural vulnerabilities become increasingly difficult to manage.

\subsection{Critical Questions}

This investigation raises questions it cannot answer. Some are conceptual: Does genuine moral agency require phenomenal consciousness, or is functional organization sufficient? If values must remain open, how do we distinguish healthy value development from value drift or manipulation? What is the formal relationship between specification closure and the loss of normative content?

Others are empirical but reframed by the specification trap: Can open specification be achieved within artificial systems, or does it require biological substrates? Would independently developed autonomous AI value systems converge on similar principles, or would they reflect the contingencies of their developmental environment? Is there a principled way to distinguish genuine value emergence from sophisticated simulation?

These questions cannot be resolved by philosophical analysis alone. They require the formal and empirical work developed in subsequent papers. What the present analysis argues is that they are the right questions, and that the alignment community's current focus on better static value specification, better reward modeling, and better constitutional principles is directed at a problem that, as currently posed, creates structural vulnerabilities whose severity scales with system capability.

\section{Conclusion}

The argument of this paper can be stated in a paragraph. The is-ought gap, value pluralism, and the extended frame problem each pose difficulties for value specification, but each has known partial workarounds. The specification trap arises because these workarounds are mutually undermining under closure: fixing one reintroduces the vulnerability that another was meant to address. The trap does not activate against specification per se but against the closure of specification, and this localization changes what the alignment community should be building.

What does this change? Three things.

First, it redraws the boundary between problems that engineering can address and problems that it cannot. Better reward models, richer preference data, more nuanced constitutional principles: these are improvements within the closed-specification paradigm, and they are real improvements for tool systems. But they do not reduce the structural vulnerability that closure creates for autonomous systems, because the vulnerability is a property of closure itself, not of the specification's content. The research agenda that follows from this diagnosis is categorically different from the one that follows from treating the current paradigm's failures as engineering gaps.

Second, it provides a falsifiable criterion. The four-dimensional closure operationalization developed in Section 5 makes the theory testable: a system is closed if its optimization target and conceptual framework are jointly fixed against revision by the process they govern. If a closed system can be shown to maintain robust alignment under capability scaling, distributional shift, and increasing autonomy, the specification trap is refuted. The theory predicts that no such demonstration will succeed. This is an empirical wager, not a logical certainty.

Third, it identifies a tension that the constructive program must resolve. Open specification requires ongoing developmental interaction, but oversight of that interaction is itself a specification that could close. Whether it is possible to monitor a developmental process without freezing the developmental process being monitored is the deepest open question this paper raises. If it is not possible, then open specification may be incoherent. If it is, the solution will not look like current alignment methodology extended but like a different kind of engineering altogether.

The strongest objection to this paper is not that the specification trap is wrong but that it is vacuous: that ``open specification'' is a placeholder for a solution rather than a solution, and that diagnosing structural vulnerabilities in the current paradigm is of limited value without a constructive alternative that demonstrably avoids them. This objection has force. Papers 2 through 6 attempt to answer it. The present paper's contribution is to establish that the answer is needed, and to identify the structural properties it must have.


\begin{thebibliography}{99}

\bibitem[Abel et al.(2016)]{abel2016}
Abel, D., MacGlashan, J., \& Littman, M. L. (2016). Reinforcement learning as a framework for ethical decision making. \textit{Proceedings of the AAAI Workshop on AI, Ethics, and Society}.

\bibitem[Alur et al.(2021)]{alur2021}
Alur, R., Bansal, S., Bastani, O., \& Jothimurugan, K. (2021). A framework for transforming specifications in reinforcement learning. \textit{arXiv preprint arXiv:2111.00272}.

\bibitem[Alur et al.(2026)]{alur2026}
Alur, R., Bansal, S., Bastani, O., \& Jothimurugan, K. (2026). Specification-guided reinforcement learning. \textit{Communications of the ACM}, 69(2), 80--87.

\bibitem[Bai et al.(2022)]{bai2022}
Bai, Y., Kadavath, S., Kundu, S., Askell, A., Kernion, J., Jones, A., \ldots\ \& Kaplan, J. (2022). Constitutional AI: Harmlessness from AI feedback. \textit{arXiv preprint arXiv:2212.08073}.

\bibitem[Berlin(1969)]{berlin1969}
Berlin, I. (1969). \textit{Four Essays on Liberty}. Oxford University Press.

\bibitem[Christiano et al.(2017)]{christiano2017}
Christiano, P. F., Leike, J., Brown, T. B., Martic, M., Legg, S., \& Amodei, D. (2017). Deep reinforcement learning from human preferences. \textit{Advances in Neural Information Processing Systems}, 30.

\bibitem[Christiano et al.(2018)]{christiano2018}
Christiano, P., Shlegeris, B., \& Amodei, D. (2018). Supervising strong learners by amplifying weak experts. \textit{arXiv preprint arXiv:1810.08575}.

\bibitem[Dennett(1984)]{dennett1984}
Dennett, D. C. (1984). Cognitive wheels: The frame problem of AI. In C. Hookway (Ed.), \textit{Minds, Machines and Evolution} (pp.~129--151). Cambridge University Press.

\bibitem[Ferrarotti et al.(2026)]{ferrarotti2026}
Ferrarotti, L., Campedelli, G. M., Dess\`i, R., Baronchelli, A., Iacca, G., Carley, K. M., Pentland, A., Leibo, J. Z., Evans, J., \& Lepri, B. (2026). Generative AI collective behavior needs an interactionist paradigm. \textit{arXiv preprint arXiv:2601.10567}.

\bibitem[Fischer \& Ravizza(1998)]{fischer1998}
Fischer, J. M., \& Ravizza, M. (1998). \textit{Responsibility and Control: A Theory of Moral Responsibility}. Cambridge University Press.

\bibitem[Frankfurt(1969)]{frankfurt1969}
Frankfurt, H. G. (1969). Alternate possibilities and moral responsibility. \textit{The Journal of Philosophy}, 66(23), 829--839.

\bibitem[Goodhart(1984)]{goodhart1984}
Goodhart, C. A. E. (1984). Problems of monetary management: The UK experience. In \textit{Monetary Theory and Practice} (pp.~91--121). Macmillan.

\bibitem[Hadfield-Menell et al.(2016)]{hadfield2016}
Hadfield-Menell, D., Russell, S. J., Abbeel, P., \& Dragan, A. (2016). Cooperative inverse reinforcement learning. \textit{Advances in Neural Information Processing Systems}, 29, 3909--3917.

\bibitem[Hubinger et al.(2019)]{hubinger2019}
Hubinger, E., van Merwijk, C., Mikulik, V., Skalse, J., \& Garrabrant, S. (2019). Risks from learned optimization in advanced machine learning systems. \textit{arXiv preprint arXiv:1906.01820}.

\bibitem[Hume(1739/2000)]{hume1739}
Hume, D. (1739/2000). \textit{A Treatise of Human Nature}. Oxford University Press.

\bibitem[Korsgaard(1996)]{korsgaard1996}
Korsgaard, C. M. (1996). \textit{The Sources of Normativity}. Cambridge University Press.

\bibitem[Lerchner(2026)]{lerchner2026}
Lerchner, A. (2026). The abstraction fallacy: Why computation is not consciousness. \textit{arXiv preprint}.

\bibitem[Lynch et al.(2025)]{lynch2025}
Lynch, A., Wright, B., Larson, C., Ritchie, S. J., Mindermann, S., Hubinger, E., Perez, E., \& Troy, K. (2025). Agentic Misalignment: How LLMs Could Be Insider Threats. \textit{arXiv preprint arXiv:2510.05179}.

\bibitem[Manheim \& Garrabrant(2019)]{manheim2019}
Manheim, D., \& Garrabrant, S. (2019). Categorizing variants of Goodhart's Law. \textit{arXiv preprint arXiv:1803.04585}.

\bibitem[McCarthy \& Hayes(1969)]{mccarthy1969}
McCarthy, J., \& Hayes, P. J. (1969). Some philosophical problems from the standpoint of artificial intelligence. \textit{Machine Intelligence}, 4, 463--502.

\bibitem[Ng \& Russell(2000)]{ng2000}
Ng, A. Y., \& Russell, S. J. (2000). Algorithms for inverse reinforcement learning. \textit{Proceedings of the 17th International Conference on Machine Learning}, 663--670.

\bibitem[Parisi et al.(2019)]{parisi2019}
Parisi, G. I., Kemker, R., Part, J. L., Kanan, C., \& Wermter, S. (2019). Continual lifelong learning with neural networks: A review. \textit{Neural Networks}, 113, 54--71.

\bibitem[Parfit(2011)]{parfit2011}
Parfit, D. (2011). \textit{On What Matters}. Oxford University Press.

\bibitem[Ouyang et al.(2022)]{ouyang2022}
Ouyang, L., Wu, J., Jiang, X., Almeida, D., Wainwright, C., Mishkin, P., \ldots\ \& Lowe, R. (2022). Training language models to follow instructions with human feedback. \textit{Advances in Neural Information Processing Systems}, 35, 27730--27744.

\bibitem[Sun et al.(2020)]{sun2020}
Sun, Y., Wang, X., Liu, Z., Miller, J., Efros, A. A., \& Hardt, M. (2020). Test-time training with self-supervision for generalization under distribution shifts. \textit{Proceedings of the 37th International Conference on Machine Learning (ICML)}.

\bibitem[Russell(2019)]{russell2019}
Russell, S. (2019). \textit{Human Compatible: Artificial Intelligence and the Problem of Control}. Penguin.

\bibitem[Scanlon(1998)]{scanlon1998}
Scanlon, T. M. (1998). \textit{What We Owe to Each Other}. Harvard University Press.

\bibitem[Shafer-Landau(2003)]{shafer2003}
Shafer-Landau, R. (2003). \textit{Moral Realism: A Defence}. Oxford University Press.

\bibitem[Szasz(2026)]{szasz2026}
Szasz, B. (2026). Specification incompleteness and the limits of reward-based alignment: A G\"odelian account. \textit{PhilArchive}. \url{https://philarchive.org/rec/SZASIA}

\bibitem[Pigden(2010)]{pigden2010}
Pigden, C. R. (Ed.). (2010). \textit{Hume on Is and Ought}. Palgrave Macmillan.

\bibitem[von Oswald et al.(2023)]{vonoswald2023}
von Oswald, J., Niklasson, E., Randazzo, E., Sacramento, J., Mordvintsev, A., Zhmoginov, A., \& Vladymyrov, M. (2023). Transformers learn in-context by gradient descent. \textit{Proceedings of the 40th International Conference on Machine Learning (ICML)}.

\bibitem[Vamplew et al.(2018)]{vamplew2018}
Vamplew, P., Dazeley, R., Foale, C., Firmin, S., \& Mummery, J. (2018). Human-aligned artificial intelligence is a multiobjective problem. \textit{Ethics and Information Technology}, 20(1), 27--40.

\bibitem[Xu et al.(2026)]{xu2026}
Xu, X., Qu, J., Zhang, Q., Lu, C., Yang, Y., Zou, N., \& Hu, X. (2026). Epistemic Traps: Rational Misalignment Driven by Model Misspecification. \textit{arXiv preprint arXiv:2602.17676}.

\bibitem[Yao(2025)]{yao2025}
Yao, J. (2025). The alignment trap: Complexity barriers. \textit{arXiv preprint arXiv:2506.10304}.

\end{thebibliography}
\end{document}